\pdfoutput=1

\documentclass[11pt]{article}

\usepackage{authblk}
\usepackage{ACL2023}

\usepackage{times}
\usepackage{latexsym}

\usepackage[T1]{fontenc}

\usepackage[utf8]{inputenc}

\usepackage{microtype}

\usepackage{inconsolata}
\usepackage{amsmath}
\usepackage{amssymb}
\usepackage{graphicx}

\title{Vietnamese multi-document summary using subgraph selection approach\\ VLSP 2022 AbMuSu Shared Task}

\author[1]{Huu-Thin Nguyen}
\author[1 2]{Tam Doan Thanh}
\author[1]{Cam-Van Thi Nguyen}
\affil[ ]{\{18021221, vanntc\}@vnu.edu.vn}
\affil[ ]{tamdt9@viettel.com.vn}
\affil[1]{VNU University of Engineering and Technology (VNU-UET)}
\affil[2]{Viettel Group}

\begin{document}
\maketitle
\begin{abstract}
Document summarization is a task to generate a
fluent, condensed summary for a document, and
keep important information. A cluster of documents serves as the input for multi-document summarizing (MDS), while the cluster summary serves as the output. In this paper, we focus on transforming the extractive MDS problem into subgraph selection. Approaching the problem in the form of graphs helps to capture simultaneously the relationship between sentences in the same document and between sentences in the same cluster based on exploiting the overall graph structure and selected subgraphs.
Experiments have been implemented on the Vietnamese dataset published in VLSP Evaluation Campaign 2022.
This model currently results in the top 10 participating teams reported on the ROUGH-2 $F\_1$ measure on the public test set.

\end{abstract}

\section{Introduction}
In multi-document summarization (MDS), the input is a set of documents, and the output is a summary that describes important information in a coherent and non-redundant manner.  It is a complex problem that has gained attention from the research community. In recent times, there have been significant improvements in MDS due to the availability of MDS datasets and advances in pretraining approaches.
Extractive and abstractive summarizing are two well-known methods for multi-document summarization. Abstractive summarizing methods try to succinctly summarize the substance of the texts, whereas extractive summarization systems aim to extract prominent snippets, phrases, or sections from documents.

Graphs capturing relationships between textual units are of great benefit to MDS, which can help create more concise, informative, and coherent summaries from multiple sources documents. Furthermore, graphs can be easily constructed by representing sentences or paragraph as graph nodes and edges. LexRank \cite{erkan2004lexrank} computes sentence importance based on a lexical similarity graph of sentences. Graph representations of documents such as discourse graph based on discourse relations \cite{christensen2013towards}. 

In the VLSP 2022 Evaluation Campaign, Vietnamese Abstractive multi-document summarization (AbMuSu) Shared Task \cite{abmusu22} is proposed to promote the development of research on abstractive multi-document summarization for Vietnamese text.
The Vietnamese abstractive multi-document summarization task's goal is to develop summarizing systems that could automatically generate abstractive summaries for a collection of documents on a given topic. The model produces an abstractive summary that is connected to the input, which is a collection of news items on the same topic. In this paper, we approach by switching back to the graph-based extractive problem, specifically, extracting summaries for multi-document by selecting sub-graphs based on the constructed graph. Inspired by study \cite{chen2021sgsum}, we exploit more edge weights representing the relationship of sentences in the same document and train and improve the model for Vietnamese dataset.

The remaining sections are structured as follows: Section 2 details the overall architecture of the model, Section 3 reports the experimental results, and finally, Section 4 is the conclusion.

\section{Model}
\begin{figure*}[ht!]
    \centering
    \includegraphics[width=\textwidth]{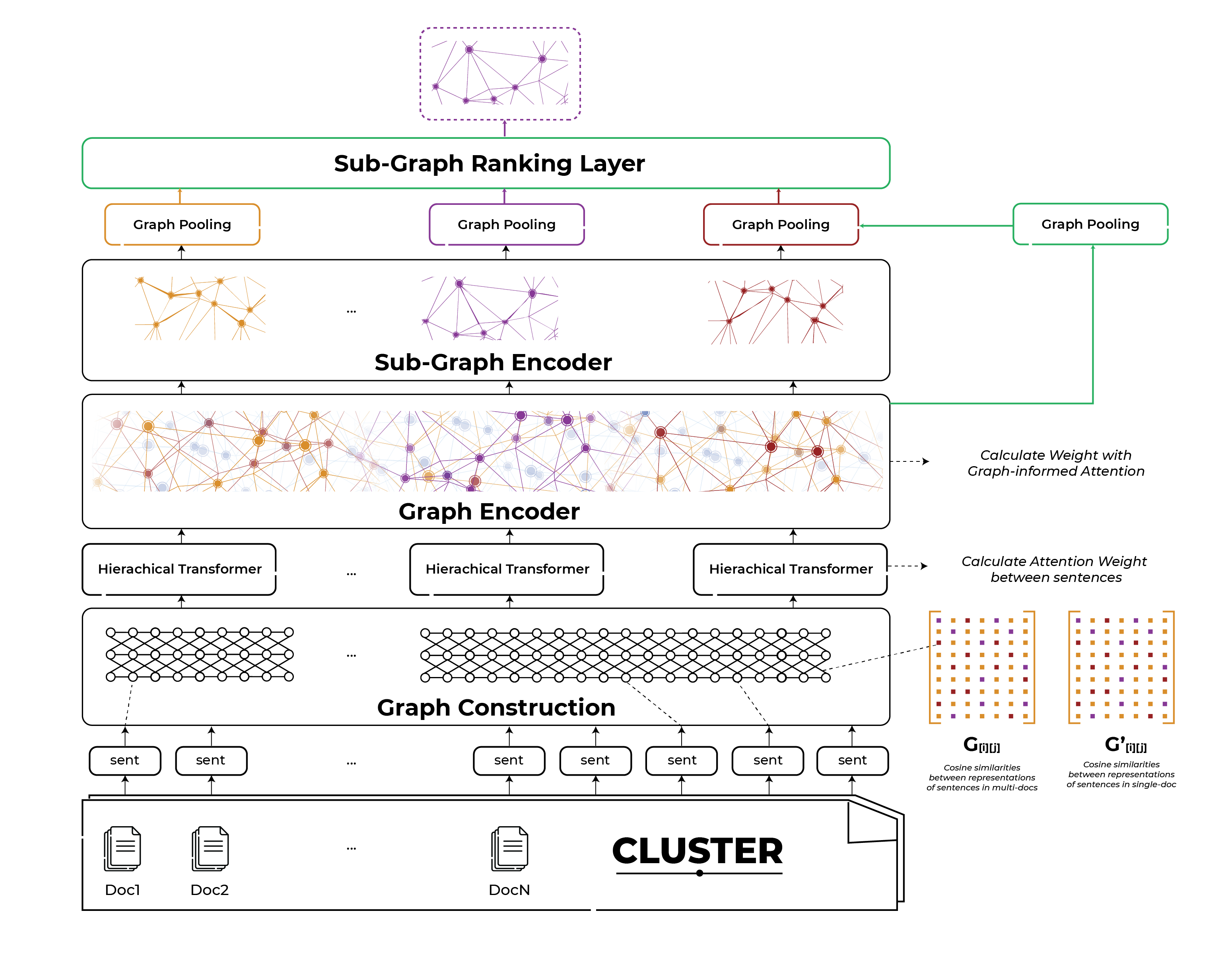}
    \caption{Model Architecture}
    \label{fig:model}
\end{figure*}

\subsection{Graph Construction}
A cluster with $N$ documents can be presented as undirected graph $\mathcal{G=(V,E)}$, where $\mathcal{V}$ denoted sentence nodes representation of each sentence in the input documents and $\mathcal{E}$ is a set of relationships between nodes. We construct graph as follow:

\subsubsection{Node}
Sentences are the basic information units and represented as nodes in the graph. 

\subsubsection{Edge}
The relations between sentences are represented as edges. We assume that all sentences in a document will be connected to each other. Therefore, any two nodes in the same documents in the same cluster are connected in the graph.

\subsubsection{Edge weight}
Let $\mathbf{G}$ denotes a graph representation matrix of the input documents, where $\mathbf{G}_{ij}$  indicates the tf-idf weights between sentence $S_i$ and $S_j$.

\subsubsection{Graph-based Multi-document Encoder}
Most of previous research simply concatenated
all documents together and treated the MDS as
a unique SDS with a longer input \cite{jin2020multi, cao2017improving}. The multi-documents will be passed through an Hierarchical Transformer, which consists of several
shared-weight single Transformer \cite{vaswani2017attention}. Each Transformer processes each tokenized document independently and outputs the sentence representations.

\subsection{Graph Encoding}
Simillar to \cite{li-etal-2020-leveraging-graph}, the pairwise relations in explicit graph representations as: 

\begin{equation}
    \alpha = Softmax(e_{ij} + \theta \times R_{ij} + \beta \times R'_{ij})
\end{equation}
where $e_{ij}$ denotes the origin self-attention weights between sentences $S_i$ and $S_j$ , $\alpha_{ij}$ denotes the adjusted weights by graph structure. The additional pairwise relation bias $R{ij}$, which is calculated as a Gaussian bias of the weights of the graph representation matrix $\mathbf{G}$, is the main component of the graph-based self-attention.

\begin{equation}
     R_{ij} = -\frac{(1-\mathbf{G}_{ij}^2)}{2\sigma^2}
\end{equation}
where $\sigma$ denotes the standard deviation that represents the influence intensity of the graph structure. 
\begin{equation}
    R'_{ij} = -\frac{(1-\mathbf{G}_{ij}^{'2})}{2\sigma^2}
\end{equation}
where $\mathbf{G}_{ij}$ represents cosine similarities between tf-idf representations
of sentences in same document. 

\subsection{Sub-graph Encoder}
Sub-graph structure can reflect the quality of candidate summaries. Having comparable nodes in a sub-graph indicates a redundant. Therefore, we use a sub-graph encoder that similar architecture to that of the graph encoder each sub-graph to be modeled. Then, in a sub-graph ranking layer, we rate each paragraph to determine which is the best as the concluding. 

\subsection{Graph Pooling}
We employ a multi-head weighted-pooling operation similar to \cite{liu-lapata-2019-hierarchical} to capture the global semantic information of source documents based on the graph representation of the documents. It uses the source graph's sentence representations as input and produces an overall representation of them (denoted as $D$), which gives global document information for both the sentence and summary selection procedures.

\subsection{Sub-graph Ranking}

We initially determine each sentence's ROUGE scores throughout the training process using the gold summary. The top-K scoring sentences are then chosen, and these are combined to create candidate summaries. Each candidate's summary's sentences make up a section of the source document graph.

The largest oracle summary (made up of source sentences) with the highest ROUGE score, which corresponds to the abstractive reference summary, is extracted using a greedy strategy \cite{nallapati2017summarunner}. Then, the oracle summary's sentences are regarded as gold summary sentences, which also form a subgraph.

Let $C^*$ signify the gold summary, and global document representation $E$, which reflects the overall meaning of source documents, should be semantically equivalent to the gold summary. 
With the gold summary, we arrange all candidate summaries in descending order of ROUGE scores. All candidate summaries are also expressed as sub-graphs using the sub-graph encoder. Naturally, the candidate pair with a higher ranking disparity should have a wider margin, which is $Loss_{pairwise}$:
\begin{equation}
\begin{aligned}
    Loss_{pairwise} = max(0, cosine(C_j, C^*) \\- cosine(C_i, C^*) +\gamma)
\end{aligned}
\end{equation}
A strong summary can capture the essence of a source document, therefore a strong sub-graph should also capture the essence of the entire graph. Particularly, the gold summary and the global document representation (D), which reflects the overall meaning of the source texts, should be semantically identical. With the largest ROUGE score matching to the abstractive reference summary, we employ a greedy technique to extract an oracle summary made up of source sentences. Then, the oracle summary's sentences are regarded as gold summary sentences, which also constitute a sub-graph.
The summary-level loss is presented as follow:
\begin{equation}
    Loss_{global} = 1 - cosine(D, C^*)
\end{equation}

$Loss_{global}$ and $Loss_{pairwise}$ compose a summary-level loss function: 
\begin{equation}
    Loss_{sum} = Loss_{pairwise} + Loss_{global}
\end{equation}

Additionally, \cite{chen2021sgsum} adopt a traditional binary cross-entropy loss between candidate sentences and oracles to learn more accurate sentence and summary representations.

\begin{equation}
    Loss_{sent} = - \sum^n_{i=1}(y_ilog(\hat{y_i}) + (1-y_i)log(1-\hat{y_i}))
\end{equation}

where a label $y_i \in {0, 1}$ indicates whether the
sentence $S_i$ should be a summary sentence. Finally, our loss can be formulated as:
\begin{equation}
    Loss =  Loss_{sent} + Loss_{sum}
\end{equation}

\section{Experiments and Results}

\subsection{Dataset}
The data includes Vietnamese news items on a variety of topics, including business, society, culture, science, and technology. It is divided into training, validation and test datasets. There are various document clusters in the datasets. Each cluster has 3-5 documents that serve as samples of the same topic. For the training and validation datasets, a manually created reference abstractive summary is provided for each cluster.
With the exception of an abstractive summary, the test set is formatted similarly to the training and validation sets. Table \ref{tab:dataset-stat} also reports the proportion of dataset.

\begin{table}[ht]
\centering
\caption{Dataset description}
\label{tab:dataset-stat}
\begin{tabular}{|c|c|c|c|}
\hline
                 & Train & Validation & Test \\ \hline
No. Cluster      & 200   & 100        & 300  \\
\# Documents           & 621   & 304        & 914  \\
Average Document & 3     & 3          & 3    \\ \hline
\end{tabular}
\end{table}

\subsection{Evaluation}
We evaluated summarization quality automatically using ROUGE-2 score \cite{lin2004rouge}. Formulas are used to determine the ROUGE-2 Recall (R), Precision (P), and $F_1$ between the predicted summary and reference summary are presented as follows:
\begin{equation}
    P= \frac{|n-grams|_{Matched}}{|n-grams|_{Predicted Summary}}
\end{equation}
\begin{equation}
    R=\frac{|n-grams|_{Matched}}{|N-grams|_{Referenced Summary}}
\end{equation}
\begin{equation}
    F_1=\frac{(2 \times P \times R)}{(P + R)} 
\end{equation}

\subsection{Training Configuration}
We use the base version of RoBERTa \cite{liu2019roberta}. to initialize our models in all experiments. The optimizer with $\mathbf{\beta}_{1}=0.9$ and $\mathbf{\beta}_{2}=0.999$, and learning rate is 0.03. Gradient clipping with maximum gradient norm 2.0 is also utilized during training. All models are trained on Google Colab Pro (1 GPU - Tesla K80) for 5 epochs. We apply dropout with probability of 0.1 before all linear layers. The number of hidden unit is set as 256, the fee feed forward hidden size is 1,024, and the number of heads is 8. The number of transformer encoding layers and graph encoding layers are set as 6 and 2, respectively. During inference we select several salient candidate nodes to build up sub-graphs. And the number of nodes in a sub-graph is determined by the average number of sentences in the gold summary. We set number of candidate nodes and sub-graph nodes as 10/9.
\subsection{Results}
We evaluate our models on AbMusu VLSP 2022 dataset. The results is presented in Table \ref{tab:results}.
Experimental results show that our model achieves the highest result of 0.28 for the rouge-2 F1 measure with $\theta=0.85$ and $\beta=0.15$.

\begin{table}[ht]
\centering
\caption{ROUGE-2 $F_1$ scores on different settings}
\label{tab:results}
\begin{tabular}{|c|c|c|}
\hline
\textbf{$\theta$} & \textbf{$\beta$} & \textbf{R-2 $F\_1$} \\ \hline
0.5            & 0.5           & 0.2397            \\
0.6            & 0.4           & 0.2552            \\
0.7            & 0.3           & 0.2573            \\
0.8            & 0.2           & 0.2811            \\
1.0            & 0.0           & 0.2715            \\
0.9            & 0.1           & 0.2733            \\
\textbf{0.85 }          & \textbf{0.15}          & \textbf{0.2823}            \\ \hline
\end{tabular}
\end{table}

\section{Conclusion and Future works}

In this paper, we have presented an approach to solving the Vietnamese multi-text summarization problem in the direction of sub-graph selection. The experimental results show quite a positive result and can be further developed in the future. In the near future, we can exploit a few more methods of recalculating edge weights in the graph besides the proposed method, or experiment with some state of the art graph neural network models.

\bibliography{sgsum}
\bibliographystyle{acl_natbib}

\end{document}